\documentclass{msdm2013}
\usepackage{subfig}
\usepackage{graphicx}
\usepackage{graphicx}
\usepackage{verbatim}
\usepackage{sidecap}
\usepackage{url}
\usepackage{bm}
\usepackage[all]{xy}
\usepackage{times}
\usepackage{subfig}
\DeclareCaptionType{copyrightbox}
\usepackage{bbding}
\usepackage[table]{xcolor}

\newcommand{\commentout}[1]{}

\newcommand{\informscommentout}[1]{}

\begin{document}

\title{A Coordinated MDP Approach to Multi-Agent Planning for Resource Allocation, with Applications to Healthcare}



%
%
%
%

\numberofauthors{3} 

\author{
%
\alignauthor
Hadi Hosseini\\
      \affaddr{David R. Cheriton School of Computer Science}\\
      \affaddr{University of Waterloo}\\
      \email{h5hosseini@uwaterloo.ca}
\alignauthor
Jesse Hoey\\
      \affaddr{David R. Cheriton School of Computer Science}\\
      \affaddr{University of Waterloo}\\
      \email{jhoey@uwaterloo.ca}
\alignauthor
Robin Cohen\\
      \affaddr{David R. Cheriton School of Computer Science}\\
      \affaddr{University of Waterloo}\\
      \email{rcohen@uwaterloo.ca}
}










\maketitle

\begin{abstract}


\begin{quote}
This paper considers a novel approach to scalable multiagent resource allocation in dynamic settings. We propose an approximate solution in which each resource consumer is represented by an independent MDP-based agent that models expected utility using an average model of its expected access to resources given only limited information about all other agents. A global auction-based mechanism is proposed for allocations based on expected regret. We assume truthful bidding and a cooperative coordination mechanism, as we are considering healthcare scenarios.
We illustrate the performance of our coordinated MDP approach against a Monte-Carlo based planning algorithm intended for large-scale applications, as well as other approaches suitable for allocating medical resources. The evaluations show that the global utility value across all consumer agents is closer to optimal when using our algorithms under certain time constraints, with low computational cost. As such, we offer a promising approach for addressing complex resource allocation problems that arise in healthcare settings.
\end{quote}

\end{abstract}








\section{\label{sec:intro}Introduction}


\commentout{
Resource planning under uncertainty in multiagent environments scales exponentially with the number of agents and resources. Therefore, in the presence of multiple agents with various possible actions and states, the size of a Markov model increases exponentially. Consequently, although multiagent MDPs (MMDP) in theory belong to the same class of complexity as MDPs (P) controlling large MDPs in multiagent systems becomes intractable due to the exponential growth in joint action and joint state space. Moreover, the decentralized control of MDPs by multiple distributed agents is shown to be NEXP-complete and computationally expensive~\cite{bernstein2002complexity}. In fact, the fully observable decentralized control (Dec-MDP) and its partially observable counterpart (Dec-POMDP) both become infeasible even when dealing with a small number of agents and resources.
}

 This paper develops an approach for allocating resources in multiagent
systems for domains where there are multiple agents and multiple tasks,
and the success of the agents carrying out tasks is dependent stochastically
on their ability to obtain a sequence of resources over time.
We are particularly interested in situations where agents must
independently optimize over their individual states, actions, and utilities,
but must also solve a complex coordination problem with other agents
in the usage of limited resources.


In particular, we are concerned with allocating resources in settings that
involve a set of $N$ {\em consumers}, each of whom requires some subset of a total of $M$ {\em resources}.  The consumers each have a measure of {\em health}\footnote{We use the term {\em health} here in a general sense to denote a single quantity over which an agent's utility function (and hence, its reward) is defined.  This can be for e.g. {\em quality} of a solution, {\em value} of an outcome,  or patient state of {\em health}.} that they are trying to optimize, and this quality is influenced stochastically by the resources they acquire and by time. 
 Further, each consumer has a resource {\em pathway} that represents the partial ordering in which they need the resources. Consumers' states evolve independently over time, and are dependent only through their need for shared resources.  Rewards are independent, and the global reward is the sum of individual consumer rewards.

We formulate this problem as a factored multiagent Markov Decision Process (MMDP) with explicit features for each consumer's state and resource utilization, and an explicit model of how each consumer's state progresses stochastically over time dependent on obtained resources.  The actions are the possible allocations of resources in each time step.  For realistic numbers of consumers and resources, however, such an MMDP has a state and action space that precludes computation of an optimal policy.  This paper addresses this problem and makes three contributions:
\begin{enumerate}
\item We develop an approximate distributed approach, where the full MMDP is broken into $N$ MDPs, one for each consumer. We call these consumer MDPs {\em agents}.  Agents model the resources they expect to obtain using a probability distribution derived from average statistics of the other agents, and compute expected regret based on this distribution and on the known dynamics of their health state.
\item We propose an iterative auction-based mechanism for real-time resource allocation based on the agents' individual expected regret values.  The iterative nature of this process ensures a reasonable allocation at minimal computational cost.
\item We demonstrate the advantages of our approach in a cooperative healthcare domain with patients seeking doctors and equipment in order to improve their health states. We present averages of simulations using randomly generated agents from a reasonable prior distribution.
We compare our coordinated MDP approach against an alternate planning algorithm intended for large-scale applications, a state-of-the-art Monte Carlo sampling based method for solving the full MMDP model known as UCT.  We also compare to two simple but realistic heuristic approaches for allocating medical resources.
\end{enumerate}

Our approach is particularly well suited to large collaborative domains that require rapid responses to resource allocation demands in time-critical domains,
and we use a healthcare scenario throughout the paper to clarify our solution.
We start by introducing the MMDP model and our distributed approach, followed by descriptions of the baseline methods we compare to.  We then develop a set of realistic models for use in simulation, and show results across a range of problem sizes.
\section{MDPs and Coordination}\label{sec:mdp-mas}

Our model is a factored MDP represented as a tuple of elements ~$\langle N, M, \tau, \mathbf{R}, \mathbf{H}, P_{T}, \Phi, A \rangle$ where $N$ is the number of consumers, $M$ the number of resources, and $\tau$ is the planning horizon.
$\mathbf{R}=\{\mathbf{R_1},\ldots,\mathbf{R_N}\}$is a finite set of resource variables, each one representing the state of a single consumer's resource utilizations, where $\mathbf{R_i}=\{R_{i1},R_{i2},\ldots,R_{iM}\}$ is a set of variables representing consumer $i$'s utilization of resource $j$. Each $R_{ij}\in\mathcal{R}$  where $\mathcal{R}$ is the set of possible resource utilizations (how much resource is being used).  We model each resource as distinct (so multiple copies of a resource are modeled separately).
$\mathbf{H}=\{H_1,\ldots,H_N\}$ is a set of $N$ variables measuring each consumer's health, 
each of which is $H_i\in\mathcal{H}$ giving the different levels of health. We use $s_i=\{\mathbf{R_i},H_i\}$ to denote the complete set of \textbf{state variables} for consumer $i$, and $S :(s_{1},...,s_{N})$ to denote the complete state for all consumers. Agent $i$ receives a reward of $\Phi_{i}(s_{i},s'_{i})$ for transition from $s_{i}$ to $s'_{i}$, thus the multiagent system's \textbf{reward function} is $\Phi(S,S') = \sum_{i}\Phi_{i}(s_{i},s'_{i})$. The \textbf{transition model} is defined as $P_{T}(S'|S,A) = \prod_{i}P_{i}(s'_{i}|s_{i},a_{i})$, which denotes the probability of reaching joint state $S'$ when in joint state $S$, and $A$ is a set of permissible \textbf{actions}, one for each resource and each consumer representing all feasible allocations of resources (so the same resource cannot be allocated to two agents simultaneously).  Resources are deterministic given the actions, and only one resource can be allocated to each consumer at a time. We assume a finite horizon undiscounted setting\footnote{This is realistic in healthcare scenarios as health states do not warrant discounting.}.


The full MDP as described is an instance of a multiagent MDP (MMDP), and will be very challenging to solve optimally for reasonable numbers of consumers and resources. The total number of states is $|S|=|\mathcal{H}|^N|\mathcal{R}|^{MN}$,
and the number of actions is $\frac{N!}{(N-M)!}$.   We will show how to compute approximate (sample-based) solutions later in this paper, but first we show our approach to distributing this large MDP into $N$ smaller MDPs, and introduce our coordination mechanism for computing approximate allocations.


\begin{figure}[htb]
\centering
\includegraphics[scale=0.24]{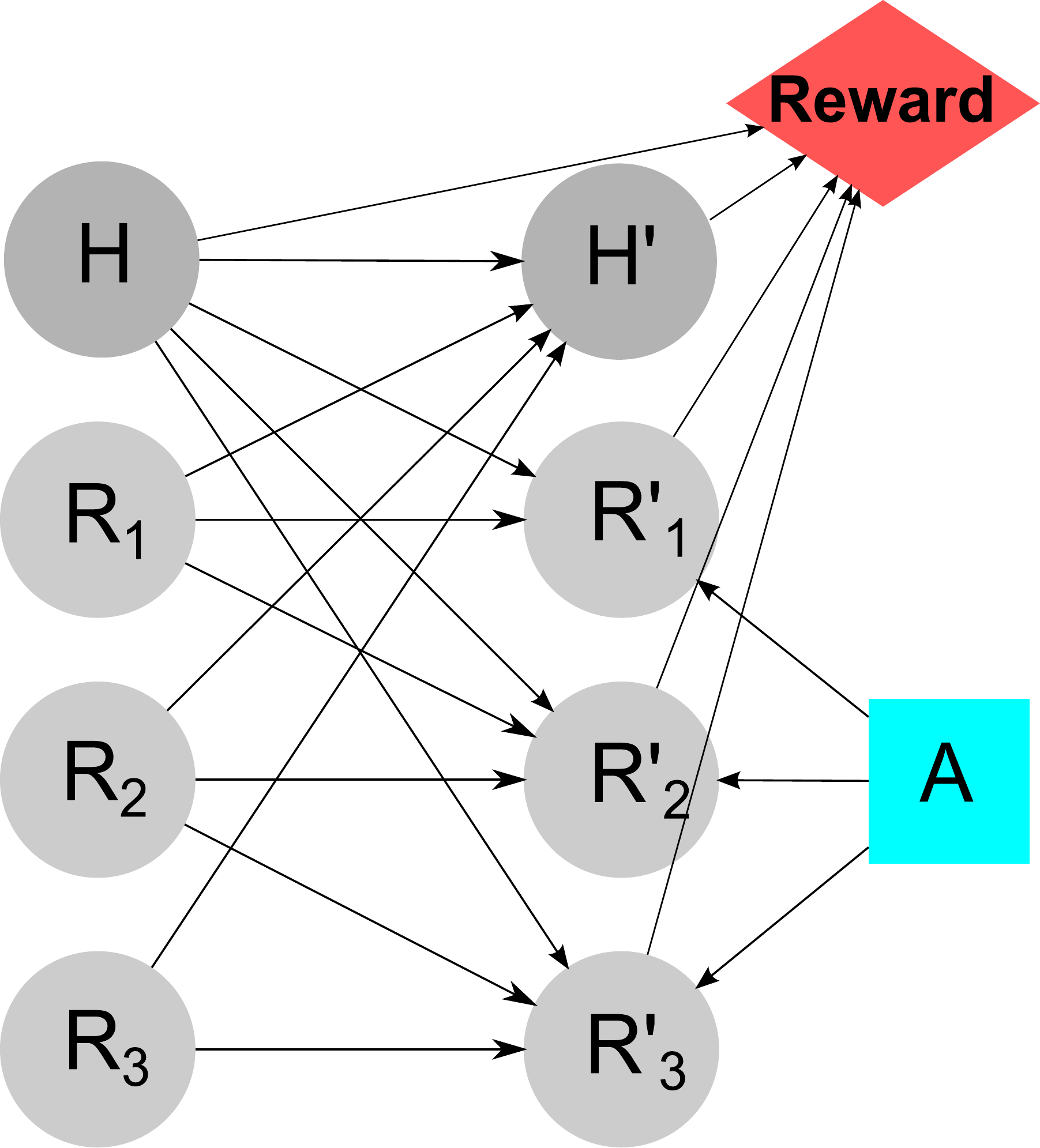}
\caption{A patient's MDP with 3 resources shown as a two time slice influence diagram}
\label{fig:ID}
\end{figure}

We treat each consumer's MDP as independent (an {\em agent}), an example of which is shown in Figure~\ref{fig:ID}.
We assume that the agent's state spaces, resource utilizations, health states, transition and reward functions are independent.
The agents are only dependent through their shared usage of resources: only feasible allocations are permitted as described above (agents can't simultaneously share resources).
Rewards are additive and each agent's actions now become {\em requests} for resources as described below.
We make two further assumptions. First, the reward function for each agent is dependent on the agent's health, H, and is set to zero by a boolean factor at the end of resource acquisition (finishing the medical pathway by receiving all required resources). 
Second, the agent health (H) is conditionally independent of the agent action given the current resources and the previous health, and the agent actions only influence the resource allocation, since the agent can only influence health indirectly by bidding for resources.  Thus, for each agent $i$, $P_{i}(\mathbf{r}',h'|\mathbf{r},h,a)$ factors as

\begin{equation}
P_{i}(\mathbf{r}',h'|\mathbf{r},h,a) = P_{i}(\mathbf{r}'|\mathbf{r},h,a) P_{i}(h'|\mathbf{r},h)
\end{equation}

where we define $\Lambda_R\equiv P_{i}(\mathbf{r}'|\mathbf{r},h,a)$ is the probability of getting the next set of resources given the current health, resources, and action, and  $\Omega_H\equiv P_{i}(h'|\mathbf{r},h)$ is a dynamic model for the agent's health rate. \commentout{For example, in the medical domain, this will depend on the condition (e.g. disease) of the person and on criticality of the health state.  While some conditions may cause a patient to deteriorate rapidly unless resources are obtained, others may have less of an effect.}  We will refer to $\Lambda_R$ as the {\em resource obtention} model and to $\Omega_H$ as the {\em health progression} model.

{\em Health progression} is a property of a particular agent's condition or task and can be estimated from global statistics about the nature of the conditions (e.g. diseases).  $\Omega_H$  must be elicited from prior knowledge about diseases and treatments, and so forms part of a {\em disease model} that we henceforth assume is pre-defined (manually, or by learning based on historical statistics).  On the other hand, the {\em resource obtention} model, $\Lambda_R$, will be dependent on the current state of the multiagent system, and is a property of how we are setting up our resource allocation mechanism and the expected regret computations of each agent.  For example, the probability of a single agent obtaining a resource will depend on (i) the number of other agents currently bidding for that resource and (ii) the agent's model of health.

\commentout{If using a single MDP for all agents (which we know to be infeasible), then resources would be deterministic given a joint allocation action.  If modeled as a decentralized POMDP, the resource probabilities would again be deterministic, but would be conditioned on the unobservable states and actions of all the other agents.  In our model, we assume that the probability of obtaining a certain resource can be approximated from a prior probability model over all other agent's actions.  This prior model can be defined based on the known distribution of diseases, and the known requirements for treatments of each disease.  We do not delve into this problem further in this paper, leaving it for future work, and simply assume we know this distribution over all other agent actions, and therefore define $\Lambda_R$ based on a set of prior probability distributions (see Experiments and Results).}

If using a single MDP for all agents as described at the start of this section, then resources would be deterministic given a joint allocation action.
If modeled as a decentralized POMDP, the resources for each consumer would be conditioned on the unobservable states and actions of all the other consumers.  In our model, we assume that the probability of obtaining a certain resource can be approximated reasonably well, either as a proior model based
on the known distribution of diseases and the known requirements for treatments of each disease, or as a learned distribution based on simulated or real experiments.



In general, we can make no assumptions about further conditional independencies in the resource allocation factor. That is, the probability of obtaining a resource $R'$ at time $t$ may depend stochastically on the set of resources at time $t-1$. However, in many domains, there may be further independencies that can be encoded in the model.  For example, in Figure~\ref{fig:ID}, resource $R_i'$ is conditionally independent of all resources $R_j$ where $j\notin \{i,i-1\}$ (for $i>1$) and for $j\notin\{i\}$ (for $i=1$), so the resources are {\em ordered} according to the (linear) medical pathway of this particular patient.
We assume that the health progression factor can be specified for each agent independently of the other agents.

A policy for each individual MDP is a function $\pi_{i}(s_{i})\mapsto A_{i}$ that gives an action for an agent to take in each state $s_{i}$. The policy can be obtained 
by computing a value function $V^{*}_{i}(s_{i})$ for each state $s_{i}\in S_{i}$, that is maximal for each state (i.e. satisfies the Bellman equation~\cite{bellman2003dynamic}). For simplicity of notation, we remove agent indices and only show the indices for resources. Thus an individual agent's value function is represented as:

\begin{equation}
V^{*}(s) = \mathop{max}\limits_{a} \gamma\sum_{s' \in S} [\Phi(s,s') + P(s'|s,a)V^{*}(s')]
\label{eqn:bellman}
\end{equation}

The policy is then given by the actions at each state that are the arguments of the maximization in Equation~\ref{eqn:bellman}.

Agents compute their expected {\em regret} for not obtaining a given resource as follows.
The expected value, $Q_{i}(h,\mathbf{r},a_i)$  for being in health state $h$ with resources $\mathbf{r}$ at time $t$,
bidding for (denoted $a_i$) and receiving resource $r_i$ at time $t+1$ is:

\[
Q_{i}\equiv \sum_{\mathbf{r}_{-i}'}\sum_{h'}P(h'|h,\mathbf{r}) V(r_i',\mathbf{r}_{-i}',h') \delta(\mathbf{r}_{-i},\mathbf{r}_{-i}')
\]
where  $\mathbf{r_{-i}}$ is the set of all resources except $r_i$ and $\delta(x,y)=1\leftrightarrow\;x=y$ and $0$ otherwise. The equivalent value for not receiving the resource, $\bar Q_{i}(h,\mathbf{r},a_i)$, is
\[\bar Q_{i}\equiv \sum_{\mathbf{r}_{-i}'}\sum_{h'}P(h'|h,\mathbf{r}) \bar V(\bar r_i',\mathbf{r_{-i}'},h') \delta(\mathbf{r}_{-i},\mathbf{r}_{-i}')
\]

Thus, the expected regret for not receiving resource $r_i$ when in $h$ with resources $\mathbf{r}$ and taking action $a_i$  is:

\begin{equation}
R_i(h,\mathbf{r},a_i) = Q_{i}-\bar Q_{i}
\label{eqn:regret}
\end{equation}

We also refer to this as the expected {\em benefit} of receiving $r_i$.   It is important for agents in this setting to consider regret (or benefit) instead of value, as two agents may value a resource the same, but one might depend on it much more (e.g. have no other option).  Value-based bids will fail to communicate this important information to the allocation mechanism.

Note that $Q$ is an optimistic estimate, since the expected value assumes the optimal policy can be followed after a single time step (which is untrue).  This myopic approximation enables us to compute on-line allocations of resources in the complete multiagent problem, as described in the next section. In the following, we will use the notion of utilitarian social welfare by aggregating the total rewards amongst all agents as an evaluation measure.

\subsection{Coordination Mechanism}
A coordination mechanism must aim to respect the health needs of the patients to maximize the overall utility. Each agent estimates its expected individual regret
given its estimate of future resources and health (as given by $\Lambda_R$ and $\Omega_H$). The regret values of different agents are compared globally, and an allocation is sought that minimizes the global regret. While the final allocation decisions are made greedily in the action-selection phase, the reported expected values of regret (for bidding) consider future rewards.

To implement this allocation, we use an iterative auction-like procedure, in which each consumer bids on the resource with highest regret. The highest bidder gets the resource, and all other agents bid on their next highest regret resource.  Agents can also {\em resign}, receive no resources for one time step, and try again in a future time step.


\commentout{
\subsection{Medical Domain Example}
\label{sec:meddomain}

In this model, the order of medical tasks is a key factor in allocating resources to patients. Patients have different medical needs based on their sickness and health conditions, and recovering from these health conditions requires series of medical tasks. However, these medical tasks are dependent on the results of the previous tasks (or tests).}


\subsection{Example}\label{sec:example}

Consider a simplified scenario with 4 agents and 4 resources. We are assuming that agents require all four resources and the expected benefits for receiving resources (or regrets for not receiving resources) based on their internal utility function have been calculated as illustrated in Table \ref{fig:example}. The worst-case scenario would be when all the agents have attributed higher benefits to the same resources, so that their desire to acquire resources is in the same order or preference.

\begin{table}[htb]
  \centering
  \scriptsize
  \subfloat[Worst-case]{\label{fig:worst}
\begin{tabular}{|c|c|c|c|c|}
  \hline
  Agents & $r_{1}$ & $r_{2}$ & $r_{3}$ & $r_{4}$\\
  \hline
  $a_{1}$ & *7 & 8 & 9 & {\bf 10} \\
  \hline
  $a_{2}$  & {\bf 1} & 3 & {*6} & 7\\
  \hline
  $a_{3}$  & 3 & {\bf *4} & 5 & 6\\
  \hline
  $a_{4}$  & {5} & 6 & {\bf 7} & *8\\
  \hline
\end{tabular}
}
\hspace{0.1cm}
\subfloat[Average-case]{\label{fig:average}
\begin{tabular}{|c|c|c|c|c|}
  \hline
  Agents & $r_{1}$ & $r_{2}$ & $r_{3}$ & $r_{4}$ \\
  \hline
  $a_{1}$ & 3 & 8 & *9 & {\bf 10} \\
  \hline
  $a_{2}$  & 1 & {\bf 3} & 6 & *7\\
  \hline
  $a_{3}$  & {\bf *6} & 4 & 5 & 3\\
  \hline
  $a_{4}$  & 5 & *6 & {\bf 7} & 8\\
  \hline
\end{tabular}
}
\caption{Example scenarios: 4 agents and 4 resources. *X shows the optimal allocation, while {\bf X} shows our method.}
\label{fig:example}
\end{table}

Agents first try to acquire the resource with highest benefit. In this scenario, all agents have associated the highest benefit to $r_{4}$, however, only one ($a_{1}$) would be successful in getting it. All agents who have lost the previous auction, will now bid for the resource with the second-highest benefit, and so on. In this case, agents $a_{2}$, $a_{2}$, $a_{3}$ all have attributed $r_{3}$ as their second highest.
Our auction-based method gives
a benefit of 22 (shown in {\bf bold} in Table~\ref{fig:worst}).
The optimal allocation has the benefit of 25 (one shown with * in Table~\ref{fig:worst}).


Table \ref{fig:average} shows an average-case scenario. Again we are assuming all agents require all the resources but with more diverse preferences over the set of resources.  Our method gets a benefit of $26$ compared to the optimal benefit of $28$.

\section{Baseline Solution Methods}

\subsection{Sample-Based}

We will compare our algorithm to the result of a sample-based solution on the full MMDP as described at the start of this section.
 UCT is a rollout-based Monte Carlo planning algorithm~\cite{kocsis2006bandit} where the MDP is simulated to a certain horizon many times, and the average rewards gathered are used to select the best action to take next.
To balance between exploration and exploitation, UCT chooses an action by modeling an independent multi-armed bandit problem considering the number of times the current node and its chosen child node has been visited according to the UCB1 policy~\cite{auer2002finite}.
In general, UCT can be considered as an any-time algorithm
and will converge to the optimal solution given sufficient time and memory \cite{kocsis2006bandit}.  UCT has become the gold standard for Monte-Carlo based planning in Markov decision processes~\cite{keller2012prost}.

To rollout at each state, we use a uniform random action selection from the set of permissible actions at each state. The permissible actions are the ones that do not cause any conflict over resource acquisition. 
Subsequently, the best action is then chosen based on the UCB1 policy.
The amount of time UCT uses for rollouts is the {\em timeout}, and is a parameter that we must set carefully in our experiments, as it directly impacts the value of the sample-based solution.   Although in some resource allocation settings lengthy decision periods would not have any impact on the efficiency of allocations, arguably, the time for making allocation decisions can be important in domains requiring urgent decisions such as emergency departments and environments exposed to significant change. Delayed decisions for critical patients with acute conditions in emergency departments can have huge impact on effectiveness of treatments \cite{chalfin2007impact}. Moreover, the allocation solution may become useless by the time an optimal decision is computed as a result of fluctuations in demand, and hence, requires recomputing the allocation decision.  We will compare to UCT using a number of different realistic {\em timeout} settings.

\subsection{Heuristic methods}
We use three heuristic methods. In the first, only the agent's level of criticality is considered (we call this ``sickest first''). In the second, we use the reported regret values and only run one round of the auction-based allocation (so only one agent gets a resource at each time step: the agent with the biggest regret for not getting it).
In the third, patients are treated in the order they arrive (first-come, first-served or FCFS - a traditional healthcare method).

\section{Experiments and Results}
\label{sec:experiments}

We demonstrate our approach in simulations with realistic probabilistic models of different conditions (e.g. diseases) and health and resource dynamics distributions.  The simulations use a random sampling of agent MDPs, drawn from a realistic prior distribution over these models.  It is important to note that we are not simply defining a single patient MDP, but rather our results are averages over randomly drawn MDPs: each simulated patient is different in each simulation, but drawn from the same underlying distribution.

We make three main assumptions. First, we assume that task durations are identical (e.g. it always takes one unit of time to consume each resource). The second assumption is that each agent is only able to bid on a single resource at each bidding round (but each bidding round includes a sequence of bids to determine the action for each MDP).
The third assumption is that all patients arrive at the same time.


\begin{figure*}[htb]
\centering
 \subfloat[]{\label{fig:iterative}\includegraphics[width=0.5\textwidth]{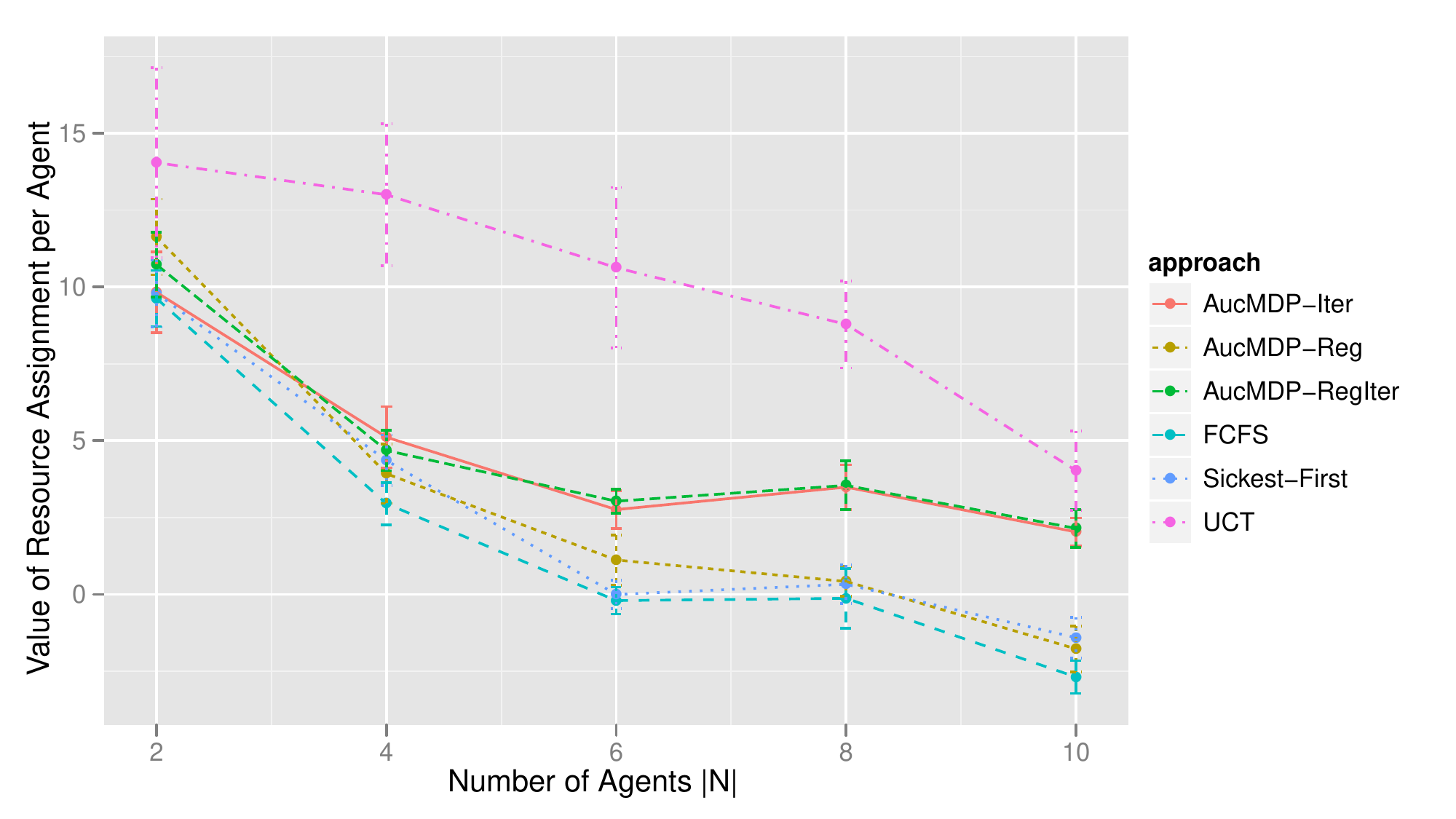}}
  \subfloat[]{\label{fig:uct_comparison}\includegraphics[width=0.5\textwidth]{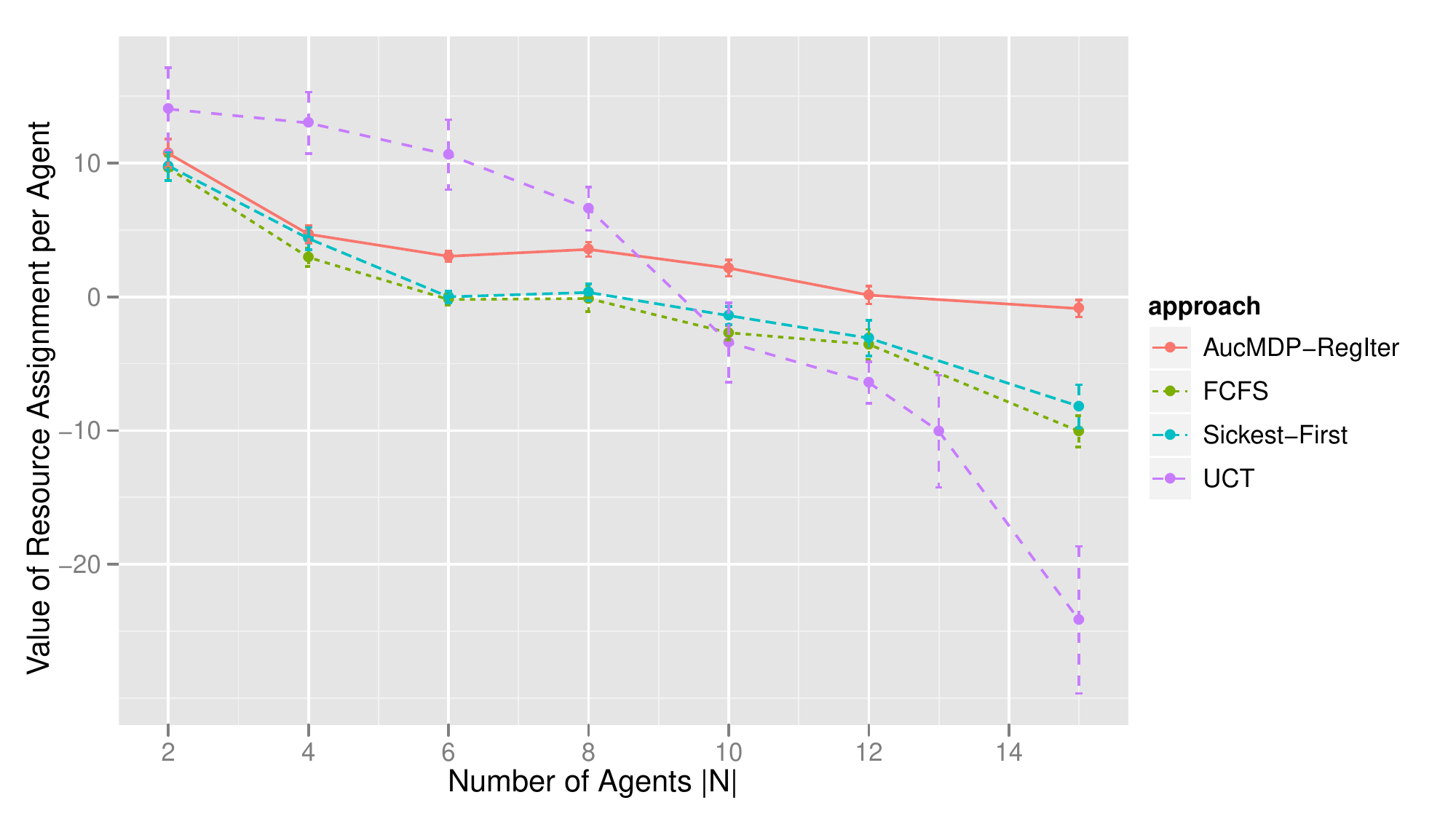}}
  \caption{Evaluation of various approaches based on expected regret (AucMDP-Reg), expected value with iteration (AucMDP-Iter), expected regret with iteration (AucMDP-RegIter), and UCT with $R=4, D=4$. (a): Timeout is 300 seconds, $\tau=10N$ 
(b): Timeout is 120 seconds, $\tau = 10N$}
\end{figure*}
\begin{figure*}[ht]
\centering
 \subfloat[]{\label{fig:uct_scale}\includegraphics[width=0.5\textwidth]{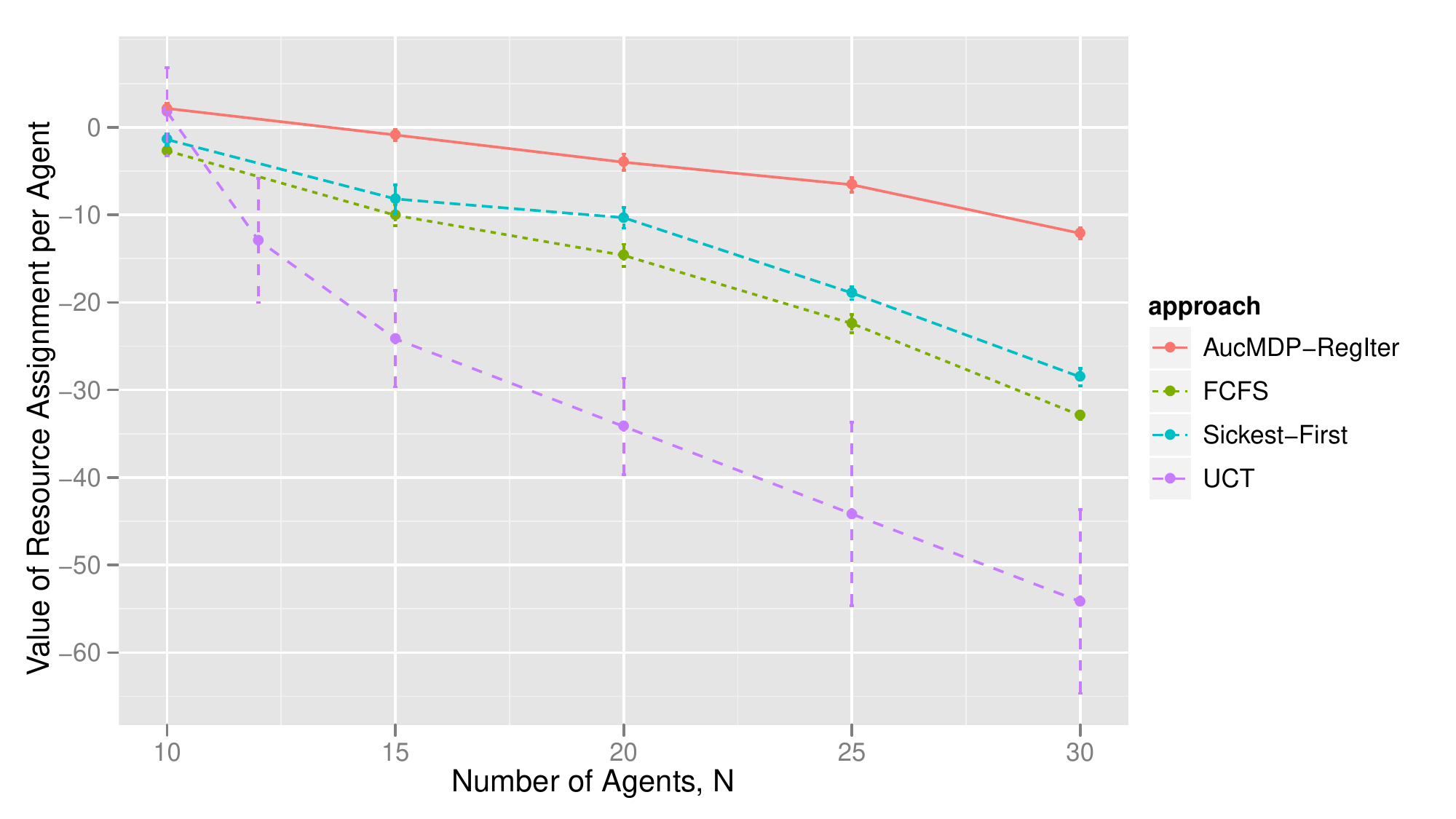}}
 \subfloat[]{\label{fig:uct_res}\includegraphics[width=0.5\textwidth]{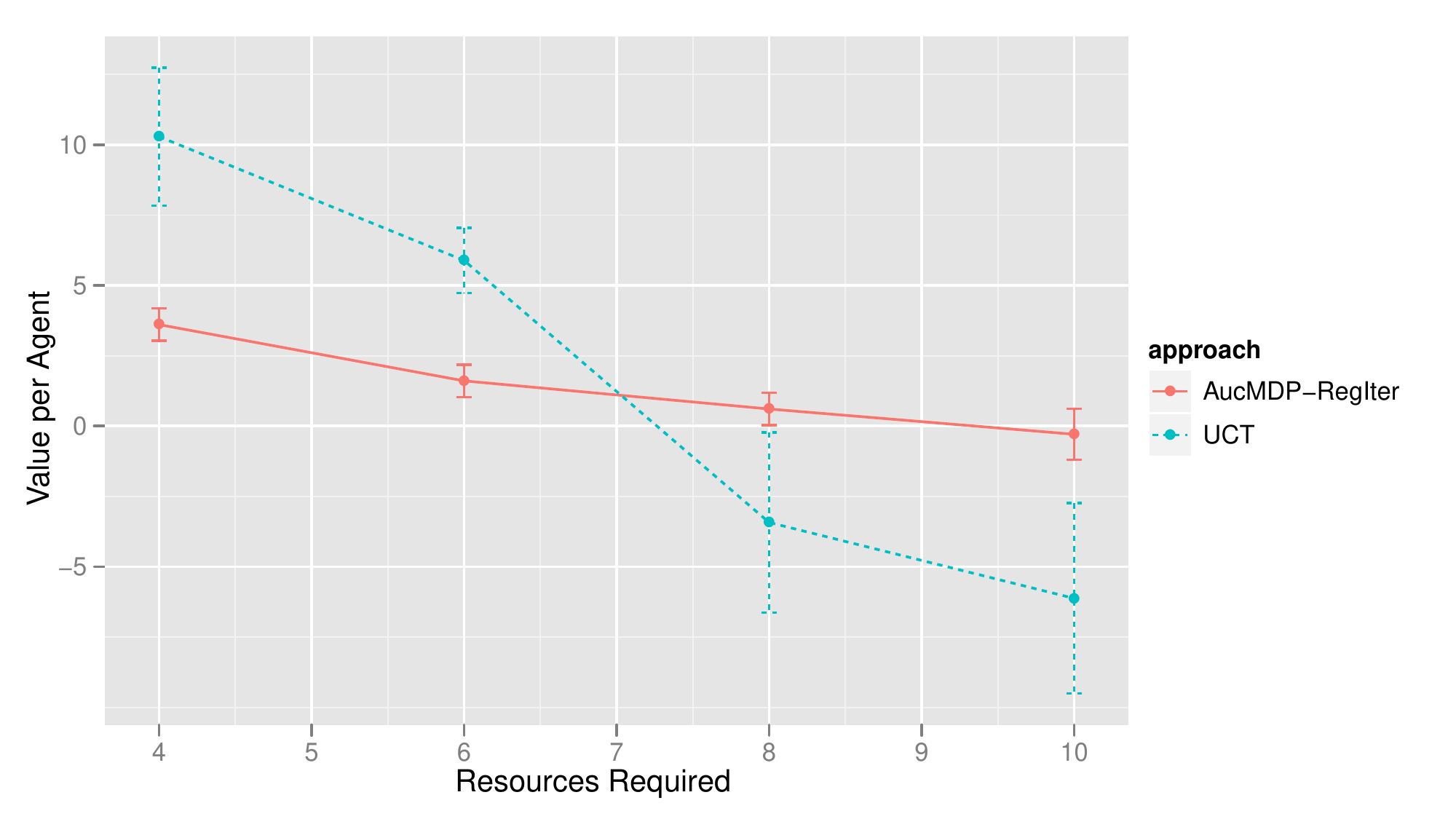}}
  \caption{(a) Scaling to 30 agents, UCT with 10mins timeout and $\tau=20$, $R=4$, $D=4$ (b) Increasing required resources (actions), UCT with 60 seconds timeout and $N=6$}
\end{figure*}

\vspace{2em}

\subsection{Agent Setup}\label{sec:agentSetup}

We assume that the health variable $H\in\{healthy,sick,critical\}$, and each resource variable $R_i\in\{have,had,need\}$.   Patients all start (enter the hospital) with $H=sick$ and, depending on the resources they acquire, their health state improves to healthy or degrades to the critical condition. We further define a function to encode the states of the health variables as $\nu(h)=\{0,1,2\}$ for $h=\{healthy, sick, critical\}$. We assume that there are $D$ possible conditions (diseases), each with a {\em criticality level}, a real number $c_d\in[1,2]$ with $c_d = 2$ being the most critical disease (makes the patient become sicker faster).

We first assume a multinomial distribution  over the $D$ conditions drawn from a set $\mathcal{D}$, such that each patient has condition $d\in\mathcal{D}$ with probability $\phi_d(d)$.   In the following, we assume conditions to be evenly distributed: $\phi_d(d)=1/|\mathcal{D}|$, although in practice this distribution would reflect the current condition distribution in the population, community or hospital. Each condition has a {\em condition profile} that specifies a set of resources in a specific order that is derived from the clinical practice guidelines or the {\em medical pathway}, a distribution over {\em health state progression} models, $\Omega_H$,
and a distribution over {\em resource obtention} models, $\Lambda_R$.

The medical pathway can be specified either within the $\Omega_H$ (by making any set of $\mathbf{r}$ not on the pathway lead to non-progression of the health state), or within $\Lambda_R$ (by making it impossible to get resource allocations outside the pathway).  We choose the latter in these experiments, but in practice the pathway may need to be specified by a combination of both, particularly if there is non-determinism in the pathways (i.e. different pathways can be chosen with different predicted outcomes).
We assume that pathways for all agents are a linear chain through the required resources for each condition.

For our experiments, we have built priors over $\Omega_H$ and $\Lambda_R$ based on our prior knowledge of the health domain. We have made these priors reasonably realistic (capture some of the main properties of this domain), and sufficiently non-specific to allow for a wide range of randomly drawn transition functions in the patient MDPs.  In practice, these priors would be elicited from experts or learned from data.

{\bf Health state progression model}: For each simulated agent, $\Omega_H$ is drawn from a Dirichlet prior distribution over the three values of $H'$ that puts more mass on the probability of healthier states (compared to the current health state) if the required resources are obtained, but more mass on the probability of sicker states if the disease is more critical.  More precisely, define $\bm{\omega}_H\sim Dir(\bm{\alpha}_H(d,\mathbf{r}))$ where $\bm{\alpha}_H$ is a triple of values over $H=\{healthy,sick,critical\}$ and $|\bm{\omega}_H| = 1$.  If all the required resources are $r=had$ in $\mathbf{r}$, then $\bm{\alpha}_H(d,\mathbf{r})=(12,4c_d,2c_d)$. If all required resources are either $r=had$, or $r=have$, then $\bm{\alpha}_H(d,\mathbf{r})=(12,4c_d,4c_d)$. Finally, if all the resources are needed, then $\bm{\alpha}_H(d,\mathbf{r})=(4,4c_d,10c_d)$. For all the other values of $\mathbf{r}$, i.e. the ones with partial resources needed, we define $\bm{\alpha}_H(d,\mathbf{r})=(4,10c_d,10c_d)$. Now for sampling purposes, we use these Dirichlet priors as parameters of multinomial distributions to sample the progression of health state.
We have assumed similar progression of health over health states for all possible transitions based on $\bm{\omega}_H:(\omega_{H,1},\omega_{H,2},\omega_{H,3})$. Thus,
\commentout{
We have assumed similar progression of health over health states for all possible transitions: \emph{get\_healthy}, \emph{stay\_same}, and \emph{get\_sick}. Hence, $P(h'=healthy|h=normal)=P(h'=normal|h=sick)$ and $P(h'=normal|h=healthy)=P(h'=sick|h=normal)$ and if $h'=h$, then $P(h'|h)$ is equal for all health states.
Thus,}
\vspace{-0.5em}
\[
\Omega_H\equiv P(h'|h,\mathbf{r})= \left\{
  \begin{array}{l l}
(\omega_{H,1},\omega_{H,2},\omega_{H,3})& \text{if }h=sick\\
(\omega_{H,1},\omega_{H,3},\omega_{H,2})& \text{if }h=healthy\\
(\omega_{H,2},\omega_{H,1},\omega_{H,3})& \text{if }h=critical
\end{array}\right.
\]
where $\omega_{H,i}$ is the $i^{th}$ element of $\bm{\omega}_H$.

{\bf Resource obtention model}: For each simulated agent, $\Lambda_R$  is drawn from a Dirichlet prior distribution over the three values of $R'$ that puts more mass on the probability of getting a resource if it is the next in the medical pathway, and if the patient is more sick (so their regret and bids will be larger, making it more likely they will get the resource). However, the probability mass shifts towards not getting a resource as $N$ gets larger (so the more agents in the system, the less likely it is to get a resource).
 Recall from above that this model is meant to summarize the joint actions of $N$ other agents, as would have been modeled in a full dec-POMDP solution. An adequate summary is important for good performance, and while we do not claim that the following prior is optimal, we believe it to be a good representation for these simulations.  Ideally this function would be computed from the complete model directly, or learned from data.
We define $\Lambda_R\sim Dir(\bm{\alpha_r}(N,h,\mathbf{r}))$ where $\bm{\alpha_r}$ is a triple of values over $R=\{have,had,need\}$.  We define
 $\nu'(h)=(1,5,10)$ for $h=(healthy,sick,critical)$. If all resources in $\mathbf{r}$ are either $had$ or $have$, then $\bm{\alpha_r}=(10\nu'(h),\nu'(h),N)$. If the previous resource in the medical pathway is $need$, then $\bm{\alpha_r}=(\nu'(h),5\nu'(h),10N)$.  Finally, if all resources are needed, then $\bm{\alpha_r}=(\nu'(h),\nu'(h),N)$.

{\bf Reward function}: $\Phi(h,h')$ is fixed for all the agents, and rewards agents for becoming healthy, but penalizes them for staying sick or going to the critical state. More precisely:
for $h' = (healthy,sick,critical)$, $\Phi(h=healthy, h') = (10,-5,-10)$, $\Phi(h=sick, h') = (15,0,-5)$, and $\Phi(h=critical, h') = (5,0,-5)$.
Further, once an patient is {\em healthy} and has received all resources, they are discharged and receive no further reward.

\subsection{Results}

We ran each of the benchmarks on a machine with 3.4GHz QuadCore AMD and 4GB RAM available. We compare our auction-based coordinated MDP approach with (AucMDP-RegIter) and without (AucMDP-Reg) iteration using the expected regret bidding mechanism. We also compare to a version where agents only bid their expected values, not regrets (AucMDP-Iter), FCFS,  sickest-first, and sample-based (UCT).
Each simulated patient is randomly assigned a condition profile and then an MDP model with parameters randomly drawn from the Dirichlet distributions defined above is assigned.
100 trials are done for each randomly drawn set of conditions and MDPs, and this is repeated 10 times.  For the UCT results, we ran $10$ trials, also repeated $10$ times.

%

We present means and standard deviations over these simulations. We first present results with 4 total resources types and each agent requiring 4 resources based on randomly assigned condition profiles (Figure~\ref{fig:iterative}). The y-axis is the average reward per patient gathered over an entire trial.  We use a horizon that depends on the number of agents ($\tau=10N$), and 
UCT is given a 300 second timeout.
The total computation time of the complete allocations for the AucMDP approach is less than 10 seconds for problems with 10 agents, and this computation time increases linearly with the number of agents and resources (as opposed to exponential growth in the MMDP case). 
We can see that the two AucMDP iterative approaches perform similarly, and outperform the heuristic approaches for $N>6$.  UCT is given sufficient time to outperform all other approaches.

Figure~\ref{fig:uct_comparison} shows the performance of our approach in a more realistic scenario with timeout set to a maximum of 120 seconds for rollouts. Similarly, each agent requires 4 resources. When the number of agents increases to more than 8 agents, UCT underperforms compared to AucMDP, providing a policy as inferior as FCFS or sickest-first. This is mostly due to the fact that the number of possible actions grows exponentially by adding more agents, and thus, UCT requires significantly more rollouts in the action exploration phase.
Figure~\ref{fig:uct_scale} shows a further scaling to $N=30$, again showing that our AucMDP approach outperforms the other methods for the larger problems.
The number of joint actions also grows exponentially when the number of resources required by each agent is increased, since there are more individual options, but our AucMDP handles this well as a result of linear growth in the number of actions (Figure~\ref{fig:uct_res}).

\begin{figure*}[htb]
\centering
 \subfloat[]{\label{fig:DiffRes}\includegraphics[width=0.5\textwidth]{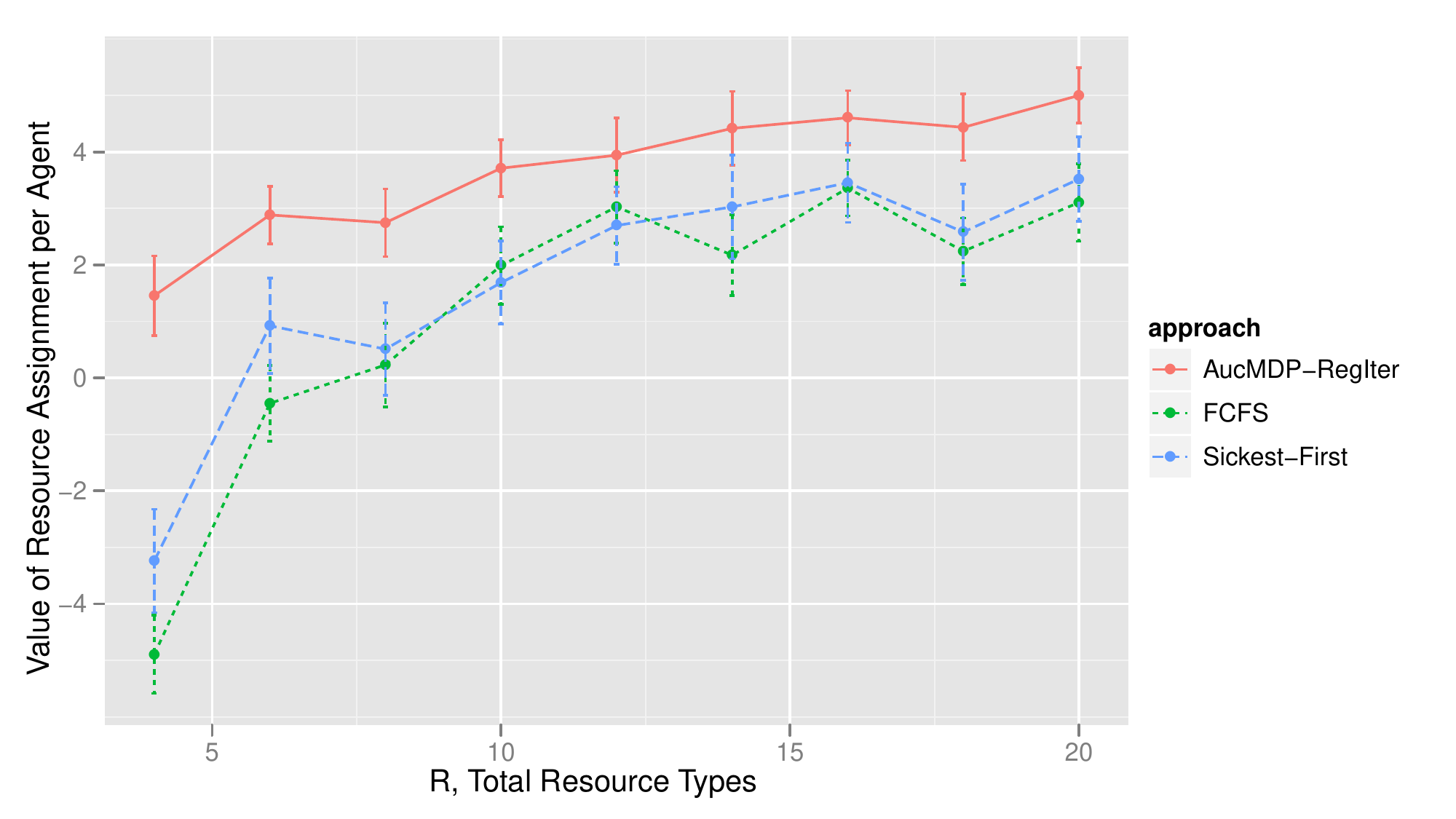}}
  \subfloat[]{\label{fig:scaling}\includegraphics[width=0.5\textwidth]{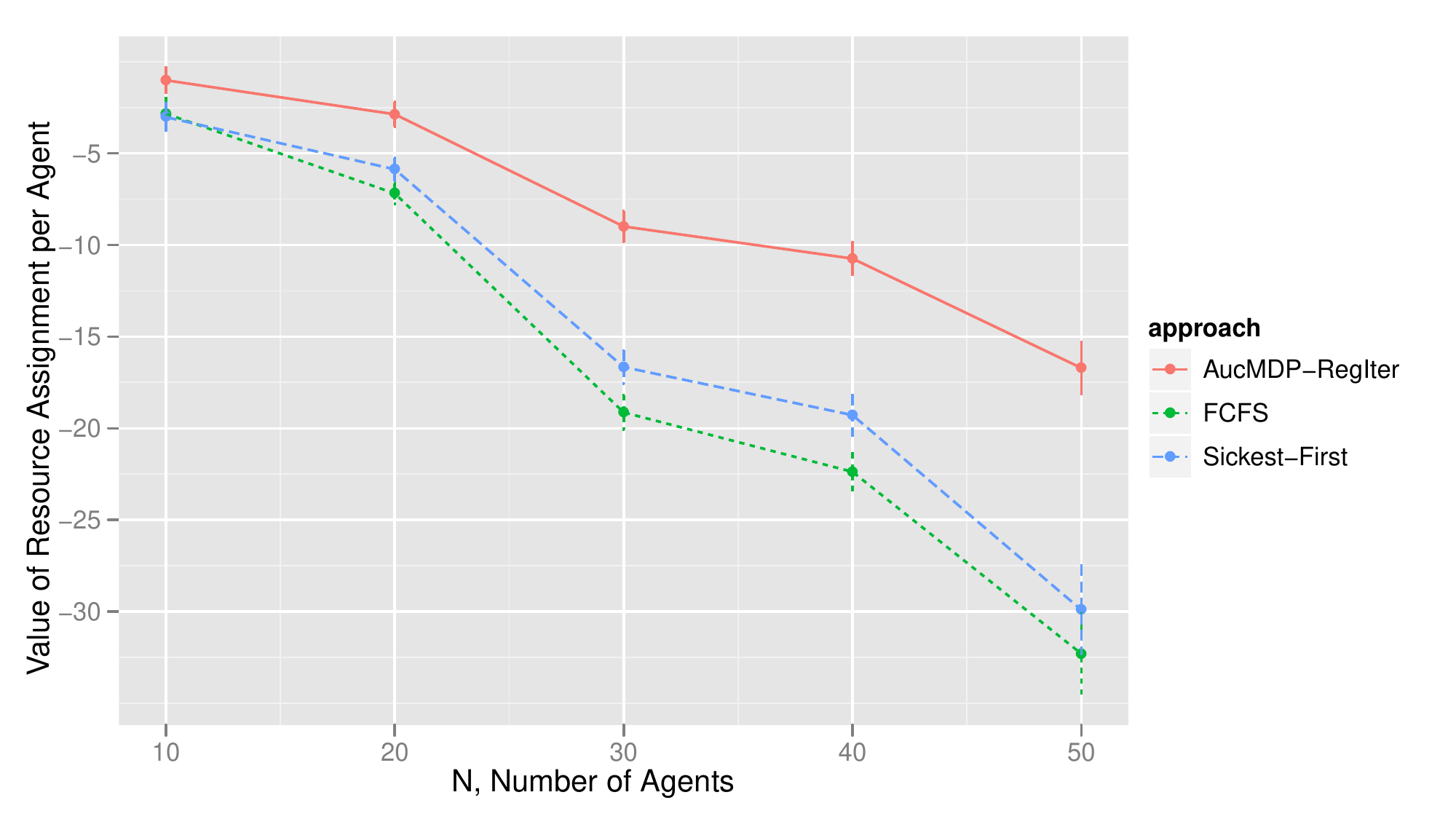}}
  \caption{(a) Varying total resource types $R=20$, $D=5$, $N=10$, more diversity in resource requirements results in fewer resource conflicts, (b) Scaling our auction-based coordination approach to $N=50$, $R=10$, $D=10$: Comparison with traditionally practiced heuristic methods in healthcare.}
\end{figure*}

As more resources are added into the system, the performance of approaches such as FCFS and sickest-first get closer to our approach because more diverse sets of resources are defined by condition profiles.
Figure \ref{fig:DiffRes} denotes that introducing more resources yields more diversity in resource requirements: the allocation problem becomes ``easier'' to solve (fewer conflicts of interest), i.e., the smaller number of resources results in harder allocation.
Figure~\ref{fig:scaling} shows results of further scaling our AucMDP approach to
50 agents each requiring 10 resources with 10 condition profiles.



%

\pagebreak
\section{Related Work and Conclusion}
Our approach to coordinating MDPs contrasts with those of multiagent MDPs~\cite{boutilier99MultiagentMDPs} and dec-MDPs~\cite{goldman2004decentralized} in finding exact solutions, which face complexity problems for large-scale problems such as ours \cite{bernstein2002complexity}. Instead, we offer an approximation method that collapses the state space of each agent down to only features that are available locally, and uses averaged effects of other agents for coordination. This is similar in spirit to~\cite{Beynier2006aaai} where effects of actions are estimated by agents (but without the central coordination, as in our work).

Our approach to resource allocation assumes additive utility independence, as in~\cite{Meuleau98}, and has state and action spaces decomposed into sets of features, with each feature relevant to only one subtask, but for cooperative settings, to maximize global utility. The use of auctions to coordinate local preferences through MDPs is also proposed in~\cite{dolgov2006resource} where individual MDPs are submitted to a central decision maker to eventually solve the winner determination problem through a mixed integer linear program (MILP). However, this model only provides one-shot allocations and is not applicable to environments with dynamic agents or resources. Multiple allocation phases are addressed in~\cite{wu2007sequential}, but the solution incurs greater communication overload with full agent preferences being modeled. Both approaches require a full preference model of all agents and their MDPs to be submitted to the auctioneer, which increases the computation effort on the side of the auctioneer for solving an MMDP and requires complicated (and often large) communication overload while raising privacy concerns. The work of~\cite{koenig2007sequential} also addresses cooperative scenarios using auctions for allocating tasks to agents with fixed types and no individual preference models. However, we employ a multi-round mechanism to assign multiple resources to dynamic agents, with expected regret dictating winner determination.

The problem of medical resource allocation is perhaps best addressed to date by~\cite{paulussen2003distributed,paulussen2006agent} which also integrates a health-based utility function to address fairness based on the severity of health states. This model does not, however, consider temporal dependency when determining allocations and our approach of considering future events provides a broader consideration of possible uncertainty. Markov decision processes have been used to model elective (non-emergency) patient scheduling in~\cite{Nunes2009}.

In all, our auction-based MDP approach addresses dynamic allocation of resources using multiagent stochastic planning, employing an auction mechanism to converge fast with low communication cost. Our experiments demonstrate effectiveness in achieving global utility, using regret, for large-scale medical applications.

Future work includes exploring auction-coordinated POMDPs~\cite{Beynier2006aaai} to estimate resource demands, and learning resource models from data. We are also interested in studying combinatorial bidding mechanisms~\cite{cramton2006introduction,rassenti1982combinatorial}, and bidding languages~\cite{nisan2000bidding} in order to optimize allocations based on richer preferences. Online mechanisms and dynamic auctions~\cite{parkes2007online} may also be of value to consider, to continue to explore changing environments.

\section{Acknowledgments}

We would like to thank the anonymous reviewers for their helpful comments.

\bibliographystyle{plain}
\bibliography{ref}
\clearpage
\end{document}